\documentclass{article}
\usepackage{amsmath}

\usepackage[preprint, nonatbib]{neurips_2024}

\usepackage[utf8]{inputenc} 
\usepackage[T1]{fontenc}    
\usepackage{hyperref}       
\usepackage{url}            
\usepackage{booktabs}       
\usepackage{amsfonts}       
\usepackage{nicefrac}       
\usepackage{microtype}      
\usepackage{xcolor}         
\usepackage{graphicx}
\usepackage{float}
\usepackage{wrapfig}

\title{Training-free Regional Prompting \\for Diffusion Transformers}

\author{
    Anthony Chen$^{1,2}$, Jianjin Xu$^{3}$, Wenzhao Zheng$^{4}$ \\ 
    \textbf{Gaole Dai}$^{1}$, \textbf{Yida Wang}$^{5}$, \textbf{Renrui Zhang}$^{6}$, \textbf{Haofan Wang}$^{2}$, \textbf{Shanghang Zhang}$^{1}$\thanks{Corresponding Author}\\
    $^{1}$Peking University,
    $^{2}$InstantX Team,
    $^{3}$Carnegie Mellon University\\
    $^{4}$UC Berkeley,
    $^{5}$Li Auto Inc.,
    $^{6}$CUHK \\
    \texttt{antonchen@pku.edu.cn}
}

\begin{document}

\maketitle

\begin{figure*}[h]
\vspace{-5mm}
  \centering
   \includegraphics[width=1\linewidth]{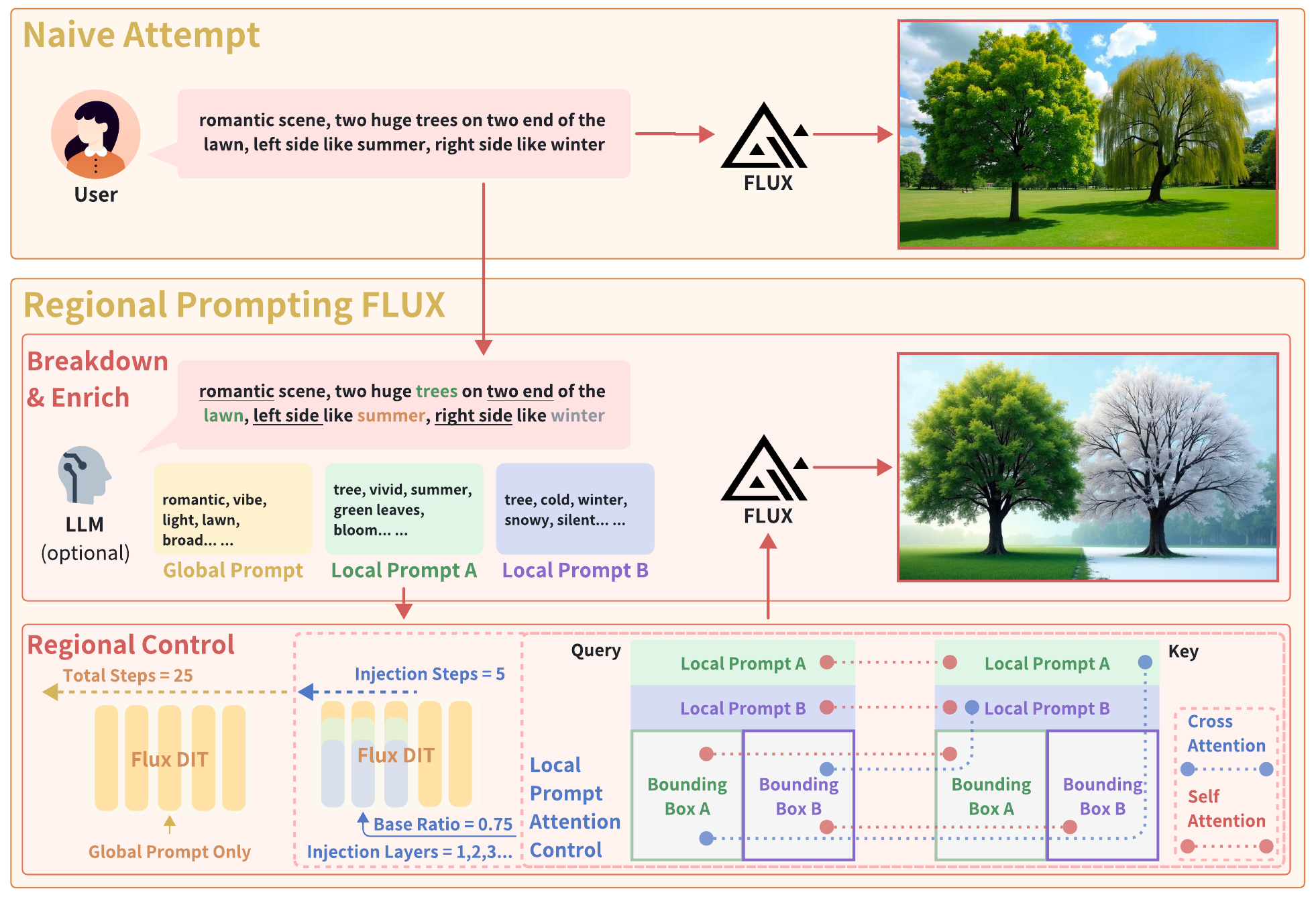}
  \vspace{-7mm}
   \caption{Overview of our method. Given user-defined or LLM-generated regional prompt-mask pairs, we can effectively achieve fine-grained compositional text-to-image generation.}
   \label{fig:intro}
  \vspace{-3mm}
\end{figure*}

\begin{abstract}
  Diffusion models have demonstrated excellent capabilities in text-to-image generation. Their semantic understanding (i.e., prompt following) ability has also been greatly improved with large language models (e.g., T5, Llama). However, existing models cannot perfectly handle long and complex text prompts, especially when the text prompts contain various objects with numerous attributes and interrelated spatial relationships. 
  While many regional prompting methods have been proposed for UNet-based models (SD1.5, SDXL), but there are still no implementations based on the recent Diffusion Transformer (DiT) architecture, such as SD3 and FLUX.1.
  In this report, we propose and implement regional prompting for FLUX.1 based on attention manipulation, which enables DiT with fined-grained compositional text-to-image generation capability in a training-free manner. Code is available at \textcolor{magenta}{https://github.com/antonioo-c/Regional-Prompting-FLUX}.
\end{abstract}

\section{Introduction}

Text-to-image models have been evolving over the past few years and have made great progress thanks to the emergence of diffusion models~\cite{rombach2022high,podell2023sdxl,esser2024scaling,flux, chen2023pixart,chen2024pixart, lin2024pixwizard}. These models come with superior visual quality, capability of generating diverse styles and promising prompt adherence given short or descriptive prompts. However, despite these remarkable advances, they still face challenges in accurately processing prompts with complex spatial layouts~\cite{yang2024mastering}. On the one hand, it is very difficult to describe the exact spatial layout through natural language, especially when the number of objects increases or precise position control is required. For example, we rely on external conditional control (such as ControlNet~\cite{zhang2023addingconditionalcontroltexttoimage}) to generate precise poses instead of describing the movements of the hand. On the other hand, although the ability of prompt adherence has been improved with the advancement of the model~\cite{flux, esser2024scaling, gao2024lumina}, when dealing with complex elements and relationships in long texts, the model still has drift problems such as confusing concepts and missing elements. Therefore, explicit spatial control is still necessary for compositional generation by now.

To tackle these challenges, people have made different attempts~\cite{flux,esser2024scaling, kolors, gao2024lumina, liu2024playground}. With the base model, research shows that the prompt following ability of text-to-image generation depends largely on the representation ability of the text encoder; that is, a larger and stronger text model can significantly enhance the prompt following ability. For example, Stable Diffusion 3/3.5~\cite{esser2024scaling} and FLUX.1~\cite{flux} additionally introduce T5-XXL~\cite{raffel2020exploring} as one of the text encoders besides of the coarse-align CLIP~\cite{radford2021learning} encoder, allowing the model to have the ability to render visual text. The Playground-3.0~\cite{liu2024playground} model further replaces the text encoder with a large language model (Lllma3~\cite{touvron2023llama}) and uses the representation of the intermediate layer instead of the global pooling representation to achieve stronger text understanding capabilities. Clearly, these advances in base models improve overall semantic understanding but still fall short of precise compositional generation to meet user needs. 

In addition to improving the base model, some recent studies~\cite{wang2024instancediffusion,yang2024mastering, li2023gligen, kim2023dense, omost} have proposed to handle compositional control by providing spatial conditions (layout/box) and training a control module as a plugin on top of the base model, or to manipulate the attention score map using region masks in a training-free manner. For example, InstanceDiffusion~\cite{wang2024instancediffusion} adds precise instance-level control via learnable UniFusion blocks to handle the additional per-instance conditioning. RPG~\cite{yang2024mastering} employs
the Multi-modal Large Language Model (MLLM)~\cite{li2024llava} as a global planner to decompose the process of generating complex images into multiple simpler generation tasks within subregions, and proposes complementary regional diffusion to enable region-wise compositional generation. DenseDiffusion~\cite{kim2023dense} and Omost~\cite{omost} develop an attention modulation method that guides objects to appear in specific regions according to layout guidance.

In this report, we are mainly inspired by Omost, but work on one of the recent diffusion transformer based models, FLUX.1-dev, which differs from previous base models mainly in its design of MMDiT structure where the prompt representation updates dynamically. We investigate training-free attention manipulation for this structure, so that it is not tied to a specific model and can be easily applied to models with similar designs. The code will be released and we hope the community can enjoy.

\section{Related Works}

\subsection{Prompt Following in Diffusion Models}

Generative models, especially diffusion-based models~\cite{rombach2022high,podell2023sdxl,esser2024scaling,flux}, have achieved remarkable progress in both image quality and semantic understanding. Latent diffusion model (LDM), originally known as Stable Diffusion 1.4/1.5~\cite{rombach2022high}, adopts a CLIP ViT-L~\cite{radford2021learning} as text encoder for text representation, and shows promising text-guided image generation at that moment. SDXL~\cite{podell2023sdxl}, as a successor, uses OpenCLIP ViT-bigG~\cite{ilharco_gabriel_2021_5143773} in combination with CLIP ViT-L~\cite{radford2021learning}, and shows that the scale and the larger text encoder help to improve the text rendering capabilities. In recent diffusion transformer based models, Pixart~\cite{chen2023pixart,chen2024pixart}, Stable Diffusion 3/3.5~\cite{esser2024scaling} and FLUX.1~\cite{flux} further use T5-XXL~\cite{raffel2020exploring} as the text encoder to handle longer and more complex prompts, and demonstrate remarkable visual text rendering capability in the absence of any explicit text prior. Hold on, scaling up text encoders seems to have immediate gains and has become a consensus, so a natural thought is, what if a large language model is used? After all, intuitively it should have better text understanding capabilities. Therefore, there are also some works have turned their attention to large language models (LLMs). Kolors~\cite{kolors} follows the structure of SDXL~\cite{podell2023sdxl}, but adopts a pre-trained General Language Model(GLM)~\cite{du2021glm} as text encoder, and exhibits an advanced comprehension of both English and Chinese, as well as its superior text rendering capabilities. Lumina-T2X~\cite{gao2024lumina} incorporate a variety of diverse text encoders with varying sizes, including CLIP~\cite{radford2021learning}, LLama~\cite{touvron2023llama, zhang2024llama}, SPHINX~\cite{lin2023sphinx}, and Phone encoders ~\cite{sun2019token}, to optimize text conditioning. But so far, all these works have only used the last hidden states of the text model or the global pooling representation. Playground V3~\cite{liu2024playground} takes a very different way, and novelly proposes to fully integrates Large Language Models (LLMs) with a novel visual structure, based on the statement that the knowledge within the LLM spans across all its layers, rather than being encapsulated by the output of any single layer. This design allows to take the hidden embedding outputs from each corresponding layer from LLM as conditioning at each
respective layer of visual model, and achieves unparalleled prompt-following and prompt-coherence abilities. In a nutshell, scaling up text encoders, cascading multiple text encoders, and leveraging multi-scale hidden representations all improve prompt following in diffusion models.

\begin{figure}[t]
  \centering
   \includegraphics[width=1.\linewidth]{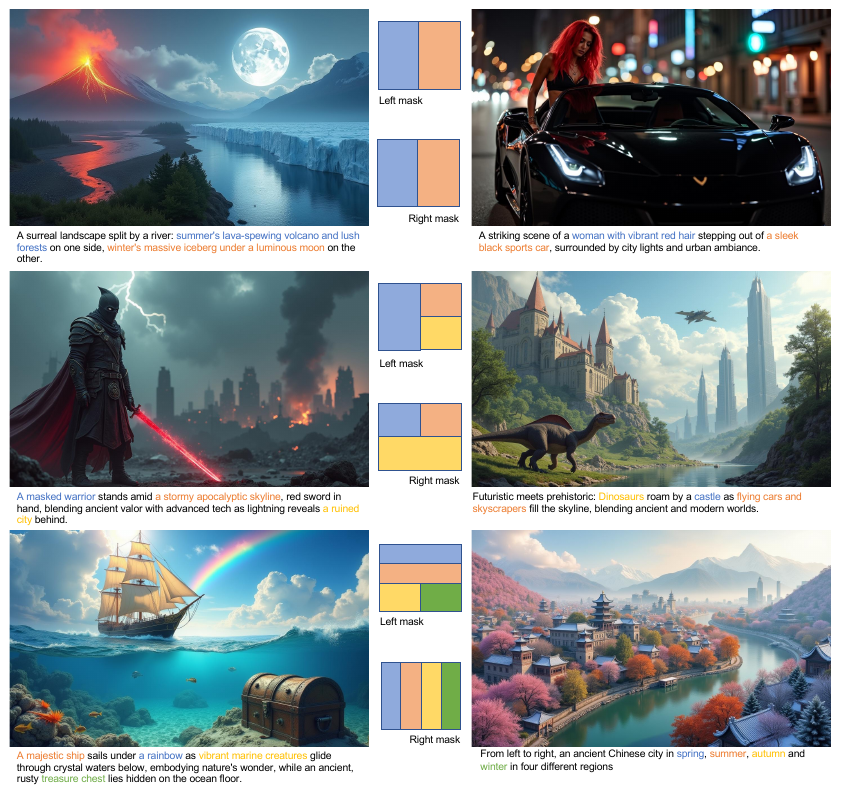}
   \vspace{-5mm}
   \caption{Main results. Simplified regional prompts are colored according to the layout mask. In practice, we input more detailed regional prompt about each region.}
   \label{fig:main}
   \vspace{-5mm}
\end{figure}

\subsection{Compositional text-to-image Generation}

Compositional generation introduces spatial conditioning to guide the image generation process with a precise layout. Although prompt adherence has been greatly improved, precise layout control that meets real-world demand is still far from enough. There have been many works~\cite{jimenez2023mixture,bar2023multidiffusion,kim2023dense,li2023gligen,wang2024instancediffusion,wang2024ms,wang2024compositional,yang2024mastering,omost} that enable region-wise compositional generation. These approaches can be roughly divided into two categories based on whether they are training-based or not. GLIGEN~\cite{li2023gligen} users Fourier embedding to encode bounding box coordinates and fuses it with corresponding text feature via a learnable projection. A new gated self-attention layer is
inserted to take in the new conditional localization information. InstanceDiffusion~\cite{wang2024instancediffusion} adds precise instance-level control via learnable UniFusion blocks to handle the additional per-instance conditioning. MS-Diffusion~\cite{wang2024ms} trains a grounding resampler to adeptly assimilate visual information and facilitates precise interactions between the image condition and the diffusion latent within the multi-subject attention layers. These works usually require training external control modules to process regional masks or bounding box. For training-free methods, Mixture of Diffusers~\cite{jimenez2023mixture} and MultiDiffusion~\cite{bar2023multidiffusion} conduct denoising steps on each region using the corresponding description, and the overall noise prediction is obtained by merging each individual noise prediction. Similarly, RPG~\cite{yang2024mastering} parallelly denoises each subregion and applies a resize-and-concatenate post-processing step to achieve high-quality compositional generation. Furthermore, DenseDiffusion~\cite{kim2023dense} and Omost~\cite{omost} develop attention modulation within cross-attention layers that guides objects to appear in specific regions according to layout guidance.

\begin{figure}[t]
  \centering
   \includegraphics[width=0.94\linewidth]{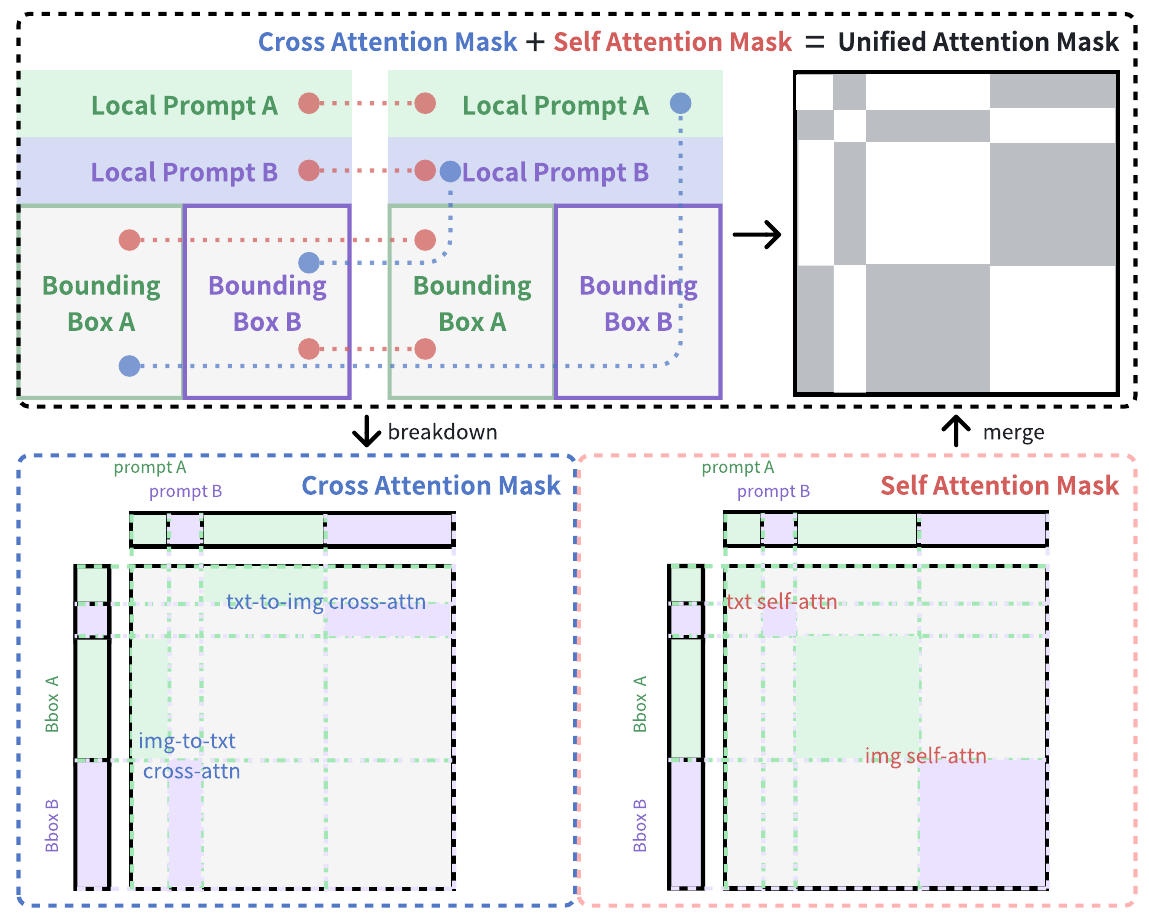}
   \vspace{-3mm}
   \caption{Illustration of our Region-Aware Attention Manipulation module. The unified self-attention in FLUX can be broken down into four parts: cross-attention from image to text, cross-attention from text to image, and self-attention between image. After calculating the attention manipulation mask, we merge them to get the overall attention mask that is later fed into the attention calculation process.}
   \label{fig:method}
   \vspace{-5mm}
\end{figure}

\section{Method}
\textbf{Task Formulation}. Our main objective is to enhance the compositional generation capabilities of current text-to-image models, enabling them to capture textual and spatial conditions in a training-free manner. To be more specific, we focus on the state-of-the-art text-to-image generation model, FLUX.1, and formally define our condition as a set of $N$ tuples and a global description. Each tuple ($c_{i}$, $m_{i}$) describes a sub-region within the image, where $c_{i}$ is a description for a region, and $m_{i}$ is a corresponding binary mask. The global description $c_{base}$ captures the overall semantic content. Given input spatial conditions, we modulate attention maps so that the layout of objects specified by $c_{i}$ can be generated within the corresponding area $m_{i}$. In practical use cases, users can automatically generate these conditions leveraging the reasoning capacity of large language models (LLM).

\textbf{Region-Aware Attention Manipulation}. We construct an attention mask $M = \{m_{ij}\} \in \mathbb{R}^{L \times L}$, where $L = L_{image} + \sum_{i=1}^N L_i$, $L_{image}$ is the length of image (e.g., $H \times W$ for a flattened 2D downsampled feature map), and $L_i$ is the length of the $i$-th regional prompt token. 

Let $X \in \mathbb{R}^{L_{image} \times D}$ represent the full image feature, where $D$ is the feature dimension. We have $N$ regional prompts $\{c_1, c_2, ...,c_N\}$ corresponding to $N$ regions $\{m_1, m_2, ...,m_N\}$ in the image.

The unified mask attention operation in MMDiT can be expressed as:
\begin{equation}
    \text{Attention}(Q, K, V, M) = \text{Softmax}\left(\frac{QK^T}{\sqrt{d_k}} \odot M\right)V,
\end{equation}

where $Q$, $K$, and $V$ are concatenated features from the image and all texts, $d_k$ is the dimension of the key vectors, and $\odot$ denotes element-wise multiplication. 

To simplify illustration, 
as in Figure~\ref{fig:method}. we can consider the unified attention operation as four parts: cross-attention from image to text, cross-attention from text to image, self-attention between image and self-attention between text. Thus, we denote our mask construction process as,

\begin{equation}
M = \begin{bmatrix}
M_{i2i} & M_{i2t} \\
M_{t2i} & M_{t2t}
\end{bmatrix}
\end{equation}

For cross-attention from image to text, similar to previous work~\cite{omost}, we apply regional masks to ensure only the image tokens within each region attend to their corresponding text, which helps maintain region-specific visual-textual associations:

\begin{equation}
    M_{i2t} = [R_1 \otimes \mathbf{1}_{1 \times |L_1|}, R_2 \otimes \mathbf{1}_{1 \times |L_2|}, ..., R_T \otimes \mathbf{1}_{1 \times |L_T|}]
\end{equation}

where $R_i \in \{0,1\}^{L_{image}}$ is the binary mask for the $i$-th region, $|L_i|$ is the number of tokens in the $i$-th text prompt, and $\otimes$ denotes the outer product.

For cross-attention from text to image, we control each query text to only attend to its corresponding region in the image, ensuring focused text-to-image interactions:
$M_{t2i} = M_{i2t}^T$

For self-attention between texts, to prevent prompt leakage and maintain the independence of each text prompt, each prompt can only attend to itself:

\begin{equation}
    M_{t2t} = \text{diag}(\mathbf{1}_{|L_1|}, \mathbf{1}_{|L_2|}, ..., \mathbf{1}_{|L_T|})
\end{equation}

For self-attention within the image, we allow attention only within each region: $M_{i2i} = \sum_{i=1}^T R_i \otimes R_i^T$. This approach maintains the integrity of region-specific information.

Our region-aware attention module ensures that each region-text pair is properly considered in the attention mechanism while maintaining the integrity of the full image feature and preventing unwanted interactions between unrelated regions and prompts. We obtain our regional latent as 

\begin{equation}
    z_{t-1}^{\text{region}} = \psi(Attention(Q_{t-1}^{region}, K_{t-1}^{region}, V_{t-1}^{region}, M^{region})
\end{equation}

where $\psi$ denotes the post-process in each transformer blocks.

To further improve the overall coherence of image compositions and ensure a harmonious transition in the boundaries of different regions, we also update base latent as

\begin{equation}
    z_{t-1}^{\text{base}} = \psi(Attention(Q_{t-1}^{base}, K_{t-1}^{base}, V_{t-1}^{base}, M^{base}))
\end{equation}

we leverage the generated latent from base prompt $c_{base}$ by combining it with the aforementioned regional latent following~\cite{yang2024mastering}.  

\begin{equation}
    z_{t-1} = \beta * z_{t-1}^{\text{base}} + (1 - \beta) * z_{t-1}^{\text{region}}
\end{equation}

The parameter $\beta$ serves as a balancing coefficient, optimizing the trade-off between aesthetic fidelity to human visual preferences and semantic alignment with the intricate textual prompt guiding the image generation process. This calibration allows for adjusting the model's output to achieve an optimal synthesis of visual appeal and prompt adherence.

\begin{figure}[H]
  \centering
   \includegraphics[width=1\linewidth]{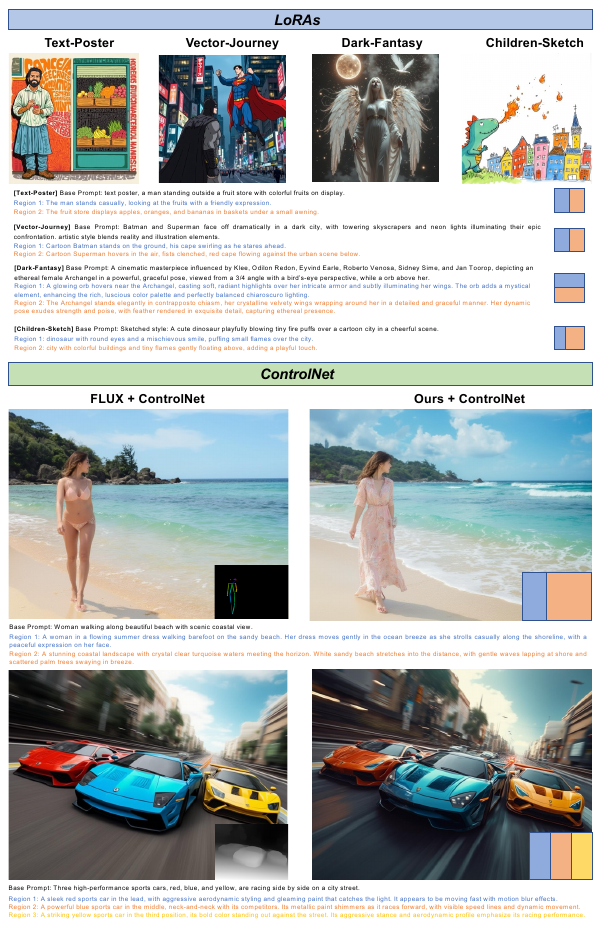}
    \vspace{-2mm}
   \caption{Results with LoRAs and ControlNet. Colored prompts and masks are provided for the regional control for each example. The control image (pose \& depth-map) for controlnet is attached within the left image. Zoom in to see in detail.}
   \label{fig:generalization}
   \vspace{-6mm}
\end{figure}

Besides, there are another two factors that matter: which denoising steps $T$ and which DiT blocks $B$, to inject regional control. Empirically, we only inject regional control in the first few denoising steps, as we find that the earlier steps in the diffusion process decide the overall layout of the image. This also helps minimize additional computation cost. For the choice of DiT blocks to inject control, we use all layers by default, but we find that the single blocks directly affect the strength of control, thus, if severe visual boundaries are observed, choosing less single layers would help.

\section{Experiments}

\begin{figure}[t]
  \centering
   \includegraphics[width=1.\linewidth]{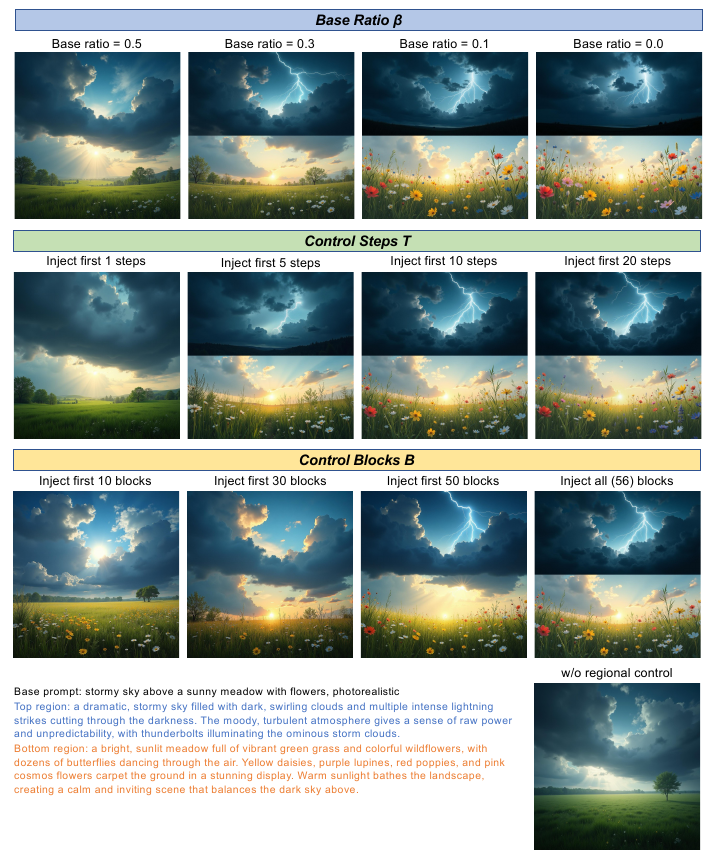}
   \vspace{-7mm}
   \caption{Ablation results with base ratio $\beta$, control steps $T$ and control blocks $B$.}
   \label{fig:ablation}
   \vspace{-5mm}
\end{figure}

\textbf{Implementation Details.}
We use FLUX.1-dev as our DiT backbone and GPT-4o as our regional prompt generator. The choice of the three factors (base ratio $\beta$, injection steps $T$, and injection blocks $B$) varies for each sample in the following figures, we manually select balanced factors for the best visual results. The experiments are conducted on a single NVIDIA A800-SXM4-80GB GPU.

\begin{wrapfigure}[19]{rt}{0.56\textwidth} 
  \centering
   \includegraphics[width=1\linewidth]{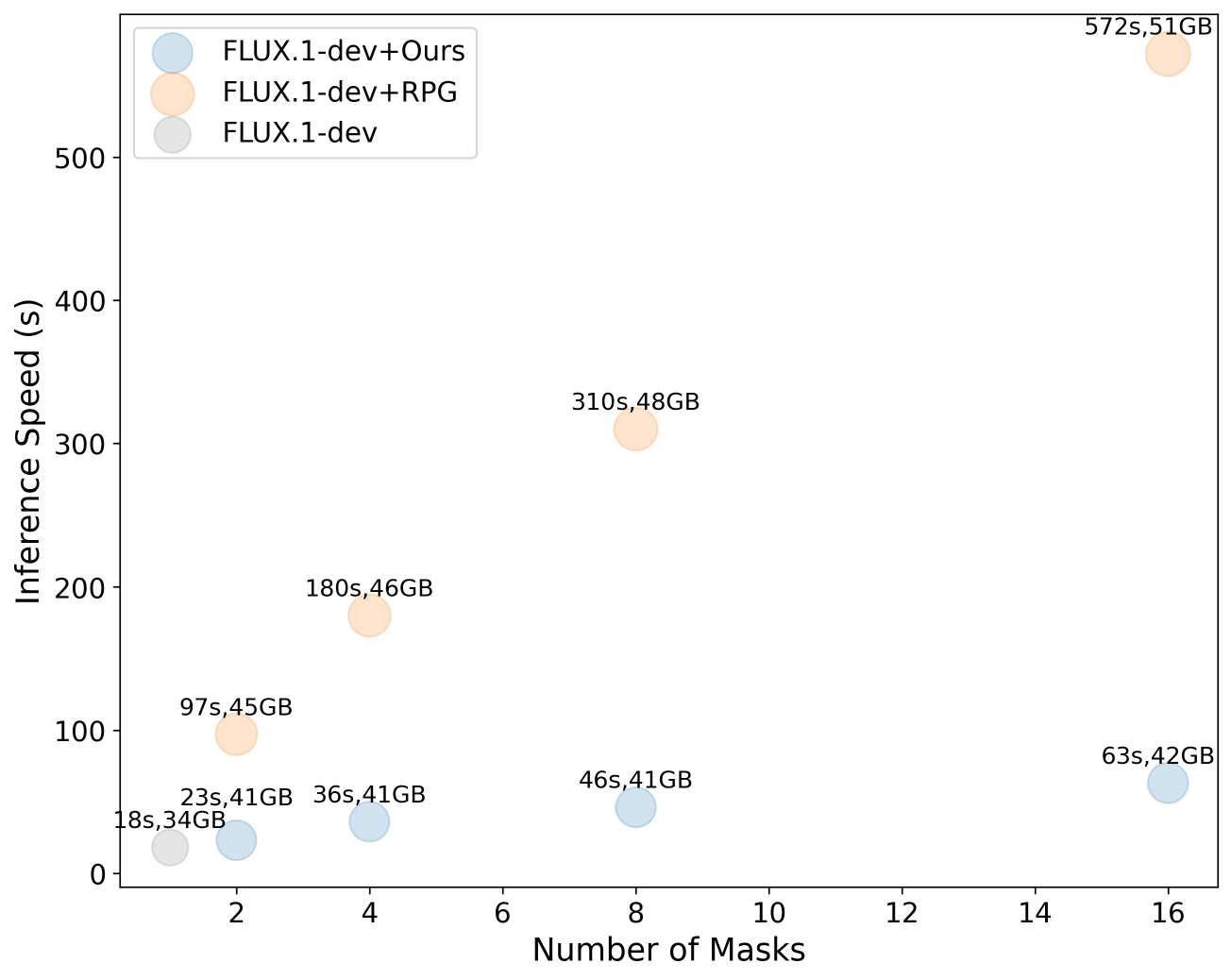}
   \vspace{-7mm}
   \caption{Inference speed and gpu memory consumption comparison with standard FLUX.1-dev, FLUX equipped with RPG-based regional control, and our method.}
   \label{fig:speed}
\end{wrapfigure}

\textbf{Main results.}
Our main results in Figure~\ref{fig:main} demonstrate our performance across various regional mask settings, highlighting its adaptability and precision in handling diverse visual prompts. As the number of regional masks increases, the model maintains strong alignment with each specified region, accurately translating the characteristics and distinctions required by the prompt into visually cohesive sections.

\textbf{Generalization Ability.}
As shown in Figure~\ref{fig:generalization}, our method can be combined with other plug-and-play modules like LoRAs~\cite{hu2021loralowrankadaptationlarge} and ControlNet~\cite{zhang2023addingconditionalcontroltexttoimage}. Provided with regional prompts, FLUX generates images that have richer details. The LoRAs and ControlNet weights are taken from public repositories by Shakerlab\footnote{https://huggingface.co/Shakker-Labs}.

\textbf{Ablation Analysis}.
Figure~\ref{fig:ablation} ablates the three factors (base ratio $\beta$, injection steps $T$, and injection blocks $B$).
We observe that while increasing regional alignment, each factor introduces trade-offs affecting image quality. Specifically, a lower base ratio, additional injection steps, and more injection blocks enhance alignment with regional prompts, creating images that better adhere to specified regions and their attributes. However, this intensified alignment often comes at the cost of severe visual boundaries between regions, disrupting the overall cohesion and aesthetic appeal. Consequently, achieving optimal image quality hinges on a balanced choice of these factors, where fine-tuning each parameter avoids harsh regional divisions while maintaining a high degree of prompt fidelity.

\textbf{Memory Consumption and Inference Speed}.
We compare our method with standard FLUX.1-dev and RPG with FLUX.1-dev. Note that this comparison is done on single NVIDIA A800-SXM4-80GB GPU. We implement RPG by generating N hidden states and later resize-and-concatenat them all together, before joining a base hidden statues in a weighted sum manner. As shown in Figure~\ref{fig:speed}, under same region masks input, we run much faster than RPG-based regional control method, when mask number reaches 16, our inference time is 9 times faster than RPG. Additionally, our GPU memory consumption is also less than RPG.

\textbf{Limitations}. 
A key limitation of our approach lies in the difficulty of tuning the factors as the number of regional masks increases. While the model is designed to handle multiple regions effectively, achieving a perfectly balanced image becomes increasingly challenging with a higher number of masks. With more regions to manage, the tuning of factors such as the base ratio, injection steps, and injection blocks requires delicate adjustment to maintain both semantic alignment with the prompt and visual cohesion across regions. This complexity often leads to trade-offs where enhancing prompt fidelity in one region may introduce unintended visual boundaries or affect the seamless integration of other regions. Consequently, as the number of masks grows, it becomes more difficult to calibrate these factors precisely to produce a cohesive and visually satisfying image.

\section{Conclusion}
We propose a training-free regional prompting method for FLUX.1, enabling fine-grained compositional generation for transformer-based models with swift and responsive image generation. Our approach enhances the prompt-following capability of FLUX.1, allowing it to handle complex, multi-regional prompts with improved semantic alignment and precise regional differentiation, all without the need for model retraining or additional data. This efficiency not only reduces the time required for image production but also streamlines users' workflows, allowing them to work more effectively. While an increased number of regional masks can make fine-tuning factors challenging for seamless outputs, this method represents a significant advancement in flexible, precise, and high-speed image generation within transformer-based frameworks.

{   
    \small
    \bibliographystyle{unsrt}
    \bibliography{neurips_2024}
}

\end{document}